\def\year{2021}\relax
\documentclass[letterpaper]{article} % DO NOT CHANGE THIS
\usepackage{aaai21}  % DO NOT CHANGE THIS
\usepackage{times}  % DO NOT CHANGE THIS
\usepackage{helvet} % DO NOT CHANGE THIS
\usepackage{courier}  % DO NOT CHANGE THIS
\usepackage[hyphens]{url}  % DO NOT CHANGE THIS
\usepackage{graphicx} % DO NOT CHANGE THIS
\usepackage{natbib}  % DO NOT CHANGE THIS AND DO NOT ADD ANY OPTIONS TO IT
\usepackage{caption} % DO NOT CHANGE THIS AND DO NOT ADD ANY OPTIONS TO IT
\usepackage{float}
\usepackage{amsmath}
\usepackage{amssymb}
\usepackage{microtype}
\usepackage{subfigure}
\usepackage{booktabs}% for professional tables
\usepackage{amsthm}
\usepackage{amsfonts}
\usepackage{arydshln}
\usepackage{multirow}
\usepackage{multicol}
\usepackage{comment}

\usepackage{tcolorbox}

\usepackage{tikz}

\usepackage{tabularx}
\usepackage{graphicx}
\usepackage{adjustbox}
\usepackage{makecell}

\usepackage{ragged2e}
\usepackage[normalem]{ulem}

\title{Outline to Story: Fine-Grained Controllable Story Generation \\ from Cascaded Events}
\author{Le Fang$^\dagger$,
    ~~Tao Zeng$^\mathsection$,
    ~~Chaochun Liu$^\mathsection$, 
    ~~Liefeng Bo$^\mathsection$, 
    ~~Wen Dong$^\dagger$,
    ~~Changyou Chen$^\dagger$\\
    \textnormal{ $^\dagger$University at Buffalo,~~
    $^\mathsection$JD Finance America Corporation, AI Lab} \\
    \textnormal{ \{lefang, wendong, changyou\}@buffalo.edu }\\ 
    \textnormal{ \{tao.zeng, chaochun.liu, liefeng.bo\}@jd.com}
 }

\date{}

\begin{document}

\maketitle

\begin{abstract}
Large-scale pretrained language models have shown thrilling generation capabilities, especially when they generate consistent long text in thousands of words with ease. However, users of these models can only control the prefix of sentences or certain global aspects of generated text. It is challenging to simultaneously achieve fine-grained controllability and preserve the state-of-the-art unconditional text generation capability. In this paper, we first propose a new task named ``Outline to Story'' (O2S) as a test bed for fine-grained controllable generation of long text, which generates a multi-paragraph story from cascaded events, {\it i.e.}, a sequence of outline events that guide subsequent paragraph generation. We then create dedicate datasets for future benchmarks, built by state-of-the-art keyword extraction techniques. Finally, we propose an extremely simple yet strong baseline method for the O2S task, which fine tunes pre-trained language models on augmented sequences of outline-story pairs with simple language modeling objective. Our method does not introduce any new parameters or perform any architecture modification, except several special tokens as delimiters to build augmented sequences. Extensive experiments on various datasets demonstrate state-of-the-art conditional story generation performance with our model, achieving better fine-grained controllability and user flexibility. Our paper is among the first ones by our knowledge to propose a model and to create datasets for the task of ``outline to story''. Our work also instantiates research interest of fine-grained controllable generation of open-domain long text, where controlling inputs are represented by short text.
\end{abstract}

\section{Introduction}
%1. 大规模模型使得free长文本生成成为可能，但是更精细的控制生成，比如段落level的控制还是有挑战。如何利用预训练模型也是一大热点。

Large-scale pretrained language models have shown thrilling generation capabilities to compose coherent and meaningful long text \cite{radford2019language,keskar2019ctrl,zellers2019defending}. However, in these models, users can only control the prefix or certain global aspects of generated text. Generation is also prone to deviate from the topic and wander to elsewhere freely. For more coherent generation, one may wish to pre-define the semantics flowing in each part, or even more explicitly control the words appearing in each paragraph. To this end, researchers face an open challenge to generate long text with fine-grained control, and simultaneously preserve the state-of-the-art unconditional text generation capabilities. 

\begin{table}[ht] \normalsize{}
\begin{tcolorbox}
\hspace{-6mm}
\begin{tabular}{l}
\textbf{Input:}
\colorbox{blue!15!white}{Prompt (Optional)}
\colorbox{blue!35!white}{Event\_1}
\colorbox{blue!35!white}{Event\_2}
\colorbox{blue!35!white}{Event\_3}
\tabularnewline
\tabularnewline
\textbf{Output:}
\colorbox{blue!35!white}{Paragraph\_1}
\colorbox{blue!35!white}{Paragraph\_2}
\colorbox{blue!35!white}{Paragraph\_3}
\tabularnewline
\end{tabular}
\end{tcolorbox}
\caption{Outline to Story. Prompt is optional but an useful addition to perform global control; each event is given to control the corresponding paragraph.}
\label{table:O2S}
\end{table}

Given the success of pre-training on a broad range of language processing tasks \cite{peters2018deep,radford2018improving,devlin2018bert}, a dominant paradigm emerges to be pre-training a transformer \cite{vaswani2017attention} based language model on a super large unlabeled corpus, and fine-tuning the model on task specific supervised data. The paradigm sheds light on transfer learning to inherit capabilities of pre-trained models. Researchers \cite{see2019massively,ziegler2019encoder} have accordingly studied the strength of massively pre-trained language models as long text generators and demonstrated their unparalleled advantages on context conditioning and generation quality. However, the need of fine-grained controllable generation is still not fulfilled, meaning that the control handle remains a single prefix or a certain global aspect such as sentiment.

%2. story生成是个很好的testbed。我们提出新的应用叫做outline to story。介绍该应用。
In this paper, we propose a new task called ``Outline to Story'' (O2S) as a test bed of fine-grained controllable generation of long text. Here, story refers to open-domain multi-paragraph long text that is self-contained and semantically sound. Given a story outline as cascaded events, {\it i.e.}, a sequence of events with one event per paragraph, where each event consists of a set of key words or phrases to appear in the corresponding paragraph, our task aims to generate a multi-paragraph story that is highly conditioned on and consistent with the outline (Table~\ref{table:O2S}). Fined-grained controllable generation is therefore performed through designated outlines. The task is challenging comparing to trivial generation of short text considering the following facts.
\begin{itemize}
    \item Firstly, we require the generated length to be much longer, i.e. from hundreds to one thousand words, to leverage the state-of-the-art text generation ability. Longer text leads to higher complexity and more flexibility in a broader space.
    \item Secondly, generated story should be highly conditioned on given cascaded events, while the events lack explicit connections and details. The generation is more constrained given more control signals than a single prompt or prefix.
	\item Lastly, a successful model on the task should not only generate well to the purpose, but also be highly flexible and controllable in generation. For instance, we humans are flexible to write stories with partial or incomplete outlines. Accordingly, an ideal model on O2S should be easily adaptable to write with only a fixed beginning event and future events on demand.
\end{itemize}

In order to tackle the proposed task, we also create some dedicate datasets. When multi-paragraph narratives are generally available, large-scale human annotated outlines are expensive and formidable. We propose to use state-of-the-art keyword extraction techniques \cite{mihalcea2004textrank,rose2010automatic} to extract a set of keywords from each paragraph and pair them with the paragraph accordingly. Note that how to get higher quality outlines is an independent and parallel research, which is out of the scope of this paper.

%3. 我们提出一种控制生成的方法，基于大规模模型，既保持文本生成能力，又实现控制。
Furthermore, we propose an extremely simple yet strong baseline method for the O2S task. Inspired by the usage of control codes and artificial tokens in recent literatures \cite{keskar2019ctrl,tsutsui2017using} to specify desired features of generated text, we use special tokens as delimiters to connect outline events with paragraphs in an interleaving manner. This leads to an augmented sequence for each outline-story pair. We fine tune pre-trained language models with the original language modeling objective on augmented sequences to learn the functionality of special tokens and co-occurrence structures between events and stories. Our method of ``fine-tuning with special tokens'' (FIST) is featured with the following facts.
\begin{itemize}
	\item Utilizing pre-trained language models enables cheap but powerful long text generation capability after transfer learning is performed.
	\item Fine-tuning on augmented sequences naturally grasps the conditioning relationship from outline to story and reserve the generation capabilities of a pre-trained model. The method does not introduce any new parameters except the latent representations of several special tokens. We also do not perform any architecture modification. Therefore, FIST has least architecture prerequisite, learning effort and model adaptation cost, while captures strong conditional dependencies.
	\item During the generation stage, the model itself is able to generate candidate outline events, and users reserve all the flexibility to modify, extend, and manipulate those events right before generating succeeding paragraphs. Users are free to start with incomplete outline. Actually it is all up to a user's practical needs to decide the level of human supervision. 
\end{itemize}

%4. 本文的概述和贡献。
To summarize, our paper is among the first ones by our knowledge to study the ``outline to story'' task along with creating new datasets for the task. It fulfills the specific need of fine-grained controllable generation, where conditions are short text. Recently, we notice a concurrent work \cite{rashkin2020plotmachines} that proposes a similar task and a dedicated architecture with memory mechanisms. Their method will be compared empirically with ours in this paper. Extensive experiments demonstrate that firstly, FIST achieves state-of-the-art conditional story generation performance; secondly, FIST has shown outstanding flexibility and controllability in the generation; lastly, FIST achieves comparable and even better metrics with its way much simpler design than \cite{rashkin2020plotmachines}. Our datasets and source code is publicly available\footnote{\url{https://github.com/fangleai/Outline2Story}}.

\section{The Task and Data}
\subsection{New Task: Outline to Story}
The ``Outline to Story'' (O2S) task aims to build strong conditional dependency between a given outline and a generated story. An outline consists of cascaded events, {\it i.e.}, a sequence of events, each corresponding to a paragraph in the story. An event is formulated as a set of keywords or key phrases that summarize the main content of a paragraph. During training and generation, an outline is expected to be self-contained and follow a latent semantic flow. The challenge of O2S task is to connect the cascaded events and fill their gap with semantically and grammatically sound details. A high quality story output is expected to strongly condition on outlines and be fruitful in content. However, it may not necessarily use all the key phrases in the outline, since a novel, creative and coherent story following the given high level semantic flow is of interest, rather than an unique particular story.

\subsection{Our Proposed Datasets}

\paragraph{Raw stories}
Our task requires datasets of paired human written outline and contentful long narratives. Unfortunately, existing public datasets rarely endow such a property. Although some tasks such as ``data to text'' \cite{chan2019stick} and ``table to text'' \cite{liu2019towards, wang2020faithful} consider similar conditional generation scenarios, the text in their datasets is generally too short and rigid. Those tasks emphasize faithful output from input data, rather than the generation capability of a model. 

Neural story generation community has contributed several candidate datasets such as $\mathtt{WritingPrompts}$ \cite{fan2018hierarchical,mao2019improving}, $\mathtt{WikiPlots}$\footnote{\url{https://github.com/markriedl/WikiPlots}}, and $\mathtt{ROCStories}$ \cite{mostafazadeh2016corpus,yao2019plan}. However, $\mathtt{ROCStories}$ only consists of 5-lines stories that are too short for our task. For the other two datasets:
\begin{itemize}
	\item $\mathtt{WritingPrompts}$ is a dedicated large-scale hierarchical story generation dataset collected from Reddit's ``WritingPromts'' forum. Based on a prompt as a rough guide or starting point, stories are multi-paragraph novels written by human users.
	\item $\mathtt{WikiPlots}$ contains plots about books, movies \emph{etc} extracted from English language Wikipedia. Each plot is paired with a short title and given as one sentence per line without paragraphing. Therefore, the first pre-processing is to segment each plot into paragraphs. We map each sentence to a fixed-length vector representation using BERT \cite{devlin2018bert,xiao2018bertservice} and allocate adjacent sentences with less cosine proximity of representations into different paragraphs. We also set a lower limit of 20 words per paragraph, since 20 words in average makes an English sentence according to linguist \cite{cutts2020oxford,campaign2004write,vincent2014sentence}.
\end{itemize}

\paragraph{Keyword Extraction}
Without human annotated outlines, we use state-of-the-art keyword extraction techniques, such as TextRank\cite{mihalcea2004textrank} and RAKE \cite{rose2010automatic} to automatically extract outline from a multi-paragraph story. Specifically, we choose RAKE\footnote{\url{https://pypi.org/project/rake-nltk/}} (Rapid Automatic Keyword Extraction), an unsupervised, domain‐independent, and language‐independent method, to extract key phrases from each paragraph. At least for English documents, RAKE is shown to be more computationally efficient than TextRank while achieving higher precision and comparable recall scores \cite{rose2010automatic}. We set minimum and maximum lengths of phrases as 1 and 4, respectively. We keep the number of extracted phrases linearly dependent on the paragraph length: it extracts one more key phrase for additional two sentences, which takes 40 words in average. These settings can be tuned as hyperparameters while the guideline is to achieve a balance between information completeness and generation creativity.

To summarize, an automatically extracted outline is expected to capture the key content and semantics flowing in a multi-paragraph story. Note that for $\mathtt{WritingPrompts}$ and $\mathtt{WikiPlots}$, we keep the given prompt or title of a story as an optional global control signal and name it ``the prompt'' in the following paper. We summarize detailed dataset statistics in Table~\ref{table:datasets}. The processed datasets will be released together with our source code for the community.

\begin{table}[t]
\begin{centering}
\begin{adjustbox}{scale={0.8}{0.8},center}
\begin{tabular}{c|c|c|c|c|c|c|c}
\toprule
\makecell{Data\\-set} & \makecell{Num.\\ stories} & \makecell{Data\\ split} & \makecell{Prompt\\ avg.\\ len.} & \makecell{Story\\ avg.\\ len.} & \makecell{Story\\ avg.\\ paragraphs} & \makecell{Event\\ avg.\\ phrases} & \makecell{Phrase\\ avg.\\ len.} \tabularnewline
\midrule
$\mathtt{WP}$ & 303 K & 90-5-5 & 25.4 & 674.5 & 6.3 & 2.8 & 2.8 \tabularnewline
$\mathtt{WI}$ & 113 K & 90-5-5 & 3.4 & 332.9 & 3.1 & 3.3 & 3.1 \tabularnewline
\bottomrule
\end{tabular}
\end{adjustbox}
\end{centering}
\caption{Statistics of two datasets. $\mathtt{WP}$: $\mathtt{WritingPrompts}$; $\mathtt{WI}$: $\mathtt{WikiPlots}$.}
\label{table:datasets}
\end{table}

\section{Methodology}

\begin{figure}[t!]
\centering
\includegraphics[scale=0.56]{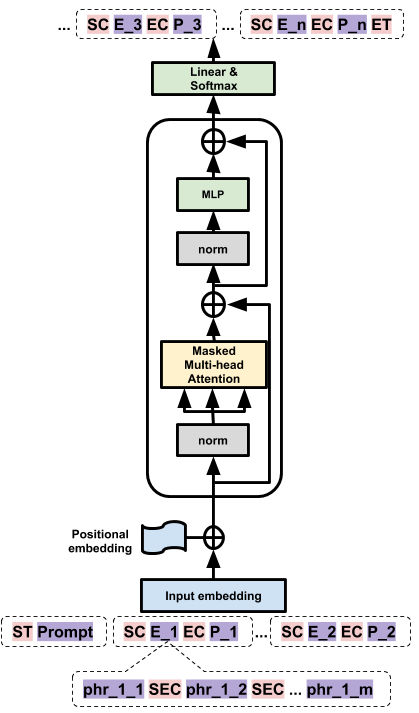}
\caption{The architecture and example of augmented sequence for language modeling in FIST. Abbreviations include \textbf{ST}: $<\!\!|\text{startoftext}|\!\!>$, \textbf{SC}: $<\!\!|\text{startofcond}|\!\!>$, \textbf{E}: an event, \textbf{SEC}: $<\!\!|\text{sepofcond}|\!\!>$, \textbf{EC}: $<\!\!|\text{endofcond}|\!\!>$, \textbf{ET}: $<\!\!|\text{endoftext}|\!\!>$, \textbf{P}: a paragraph, \textbf{phr}: a key phrase. Special tokens colored in light red are not split in tokenization and learned together with normal text.}
\label{fig:FIST}
\end{figure}

\begin{table*}[ht] \normalsize{}
\begin{tcolorbox}
\hspace{-7mm}
\begin{tabular}{l}
The interleaving manner:\tabularnewline
\begin{tikzpicture}
    \node[draw,dashed]{\colorbox{red!20!white}{ST} \colorbox{blue!35!white}{Prompt}};
\end{tikzpicture}
\begin{tikzpicture}
    \node[draw,dashed]{\colorbox{red!20!white}{SC} \colorbox{blue!35!white}{E\_1} \colorbox{red!20!white}{EC} \colorbox{blue!35!white}{P\_1}};
\end{tikzpicture}
\begin{tikzpicture}
    \node[draw,dashed]{\colorbox{red!20!white}{SC} \colorbox{blue!35!white}{E\_2} \colorbox{red!20!white}{EC} \colorbox{blue!35!white}{P\_2}};
\end{tikzpicture}
\begin{tikzpicture}
    \node[draw,dashed]{\colorbox{red!20!white}{SC} \colorbox{blue!35!white}{E\_3} \colorbox{red!20!white}{EC} \colorbox{blue!35!white}{P\_3}};
\end{tikzpicture}
$\cdots$
\begin{tikzpicture}
    \node[draw,dashed]{\colorbox{red!20!white}{SC} \colorbox{blue!35!white}{E\_n} \colorbox{red!20!white}{EC} \colorbox{blue!35!white}{P\_n} \colorbox{red!20!white}{ET}};
\end{tikzpicture} \tabularnewline
The prepending manner:\tabularnewline
\begin{tikzpicture}
    \node[draw,dashed]{\colorbox{red!20!white}{ST} \colorbox{blue!35!white}{Prompt}};
\end{tikzpicture}
\begin{tikzpicture}
    \node[draw,dashed]{\colorbox{red!20!white}{SC} \colorbox{blue!35!white}{E\_1} \colorbox{red!20!white}{EC}};
\end{tikzpicture}
\begin{tikzpicture}
    \node[draw,dashed]{\colorbox{red!20!white}{SC} \colorbox{blue!35!white}{E\_2} \colorbox{red!20!white}{EC}};
\end{tikzpicture}
\begin{tikzpicture}
    \node[draw,dashed]{\colorbox{red!20!white}{SC} \colorbox{blue!35!white}{E\_3} \colorbox{red!20!white}{EC}};
\end{tikzpicture}
...
\begin{tikzpicture}
    \node[draw,dashed]{\colorbox{red!20!white}{SC} \colorbox{blue!35!white}{E\_n} \colorbox{red!20!white}{EC}};
\end{tikzpicture}
\begin{tikzpicture}
    \node[draw,dashed]{\colorbox{blue!35!white}{P\_1} \colorbox{blue!35!white}{P\_2} \colorbox{blue!35!white}{P\_3}...\colorbox{blue!35!white}{P\_n} \colorbox{red!20!white}{ET}};
\end{tikzpicture}    \tabularnewline
\end{tabular}
\end{tcolorbox}
\caption{Two ways to build augmented sequence for each outline-story pair. Abbreviations follow that in Figure~\ref{fig:FIST}.}
\label{table:augmented_seq}
\end{table*}

At the core of our approach is the language modeling task, which is formulated as unsupervised distribution estimation of a sequence of tokens. Given a set of language examples, $(x_1, x_2, \cdots, x_k)$, each composed of variable length sequences of tokens $(s_1^k, s_2^k, \cdots, s_n^k)$, it is common to factorize the joint probabilities using the chain rule of conditional probabilities \cite{bengio2003neural}: 
\begin{align}
p(x)=\prod_{i=1}^{n}p(s_i|s_1, \cdots, s_{i-1})
\label{eq:LM_joint}
\end{align}
This decomposition allows both tractable sampling from and estimation of $p(x)$ and any conditionals of the form $p(s_{n-i}, \cdots, s_{n}|s_1, \cdots, s_{n-i-1})$. Various models are trained as distribution approximators to minimize the negative log-likelihood over a dataset $\mathcal{D}=\{x_1, x_2, \cdots, x_{\left|\mathcal{D}\right|}\}$:
\begin{align}
\mathcal{L} (\mathcal{D})=-\sum_{k=1}^{\left|\mathcal{D}\right|}\sum_{i=1}^{n}\text{log }p(s_i^k|s_{<i}^k)
\label{eq:LM_objective}
\end{align}
In recent years, large transformer-based models \cite{vaswani2017attention}, especially the GPT-2 model \cite{radford2019language}, have shown superb capabilities to estimate the conditional probabilities owing to their self-attention architectures as powerful feature extractors. Considering the nature of languages in terms of sequential ordering and inherent co-occurrence, language modeling has been a fundamental task in the field. 

In this paper, we highlight that language modeling provides a flexible way to cast various conditional distribution estimations of sequences into an unified estimation framework. A general supervised estimation task expressed as estimating $P(\text{output}|\text{input})$ is possibly solvable by language modeling on the augmented sequence of (input, output) connected by delimiters. Note that the key requirement is the sequential nature of both input and output, which is exactly the case for the ``outline to story'' task.

\subsection{Fine-tune with Special Tokens}

We adopt the common two-stage ``pre-training + fine-tuning'' paradigm. The pre-training exploits abundant unlabeled language corpus to model open-domain unconditional sequences. A continual language modeling task in fine-tuning leads to least modification on architecture and distribution space, which consequently ensures efficient model adaptation. Inspired by the usage of control codes and artificial tokens in recent literatures \cite{keskar2019ctrl,tsutsui2017using} to specify desired features of generated text, we use special tokens as delimiters to connect outline events with paragraphs. Consequently, each outline-story pair leads to an augmented sequence for language modeling in fine-tuning. Our method of ``fine-tuning with special tokens'' (FIST) naturally grasps the conditional relationship from outline to story and reserves the generation capabilities of the pre-trained model.

Take the GPT-2 \cite{radford2019language} pre-trained model as an example. Like the token  ``$<\!\!|\text{endoftext}|\!\!>$'' to indicate the end of current article and the start of next article, special tokens will not be split during tokenization. Instead, they will be learned together with normal tokens and encode certain semantic meanings in a profound way. For each outline-story pair, FIST builds an augmented sequence as following:
\begin{enumerate}
  \item For each outline event, {\it i.e.}, a set of key phrases, FIST builds an event sequence by separating key phrases with a ``$<\!\!|\text{sepofcond}|\!\!>$'' token, prepending a ``$<\!\!\!\!|\text{startofcond}|\!\!\!\!>$'' token to initiate the event, and appending a ``$<\!\!|\text{endofcond}|\!\!>$'' token to halt the event. The optional global prompt will be initiated with a ``$<\!\!|\text{startoftext}|\!\!>$'' token and put at the beginning of the first event sequence to interact with the whole content in the story.
  \item For a list of event sequences and paragraphs, FIST joins them in an interleaving manner.
\end{enumerate}
The architecture and example of augmented sequence are shown in Figure~\ref{fig:FIST}.

There are alternative ways to join a list of event sequences and paragraphs. As shown in Table~\ref{table:augmented_seq}, we advocate the best strategy in FIST to join them in an interleaving manner. Another possible way is to prepend all event sequences at the beginning and before all paragraphs. This has the advantage of exploiting all outline information starting from the very first paragraph. However, the disadvantages are also obvious. First, it weakens the paired and sequential correspondences between events and paragraphs; second, it implicitly requires the generation should have all outline events fixed before even generating the first word, hurting the generation flexibility and controllability. We will demonstrate in experiment section the benefits of FIST with an interleaving co-occurrence structure.

To summarize, special tokens are flexibly designed to accommodate various co-occurrence structures, whose representations are learned from scratch during the fine-tuning and play a significant role in the modeling of conditional dependence.

\subsection{Training and Generation}
With a pre-trained model, only a light fine-tuning on relatively small supervised data is needed. At the training stage, FIST performs language modeling on the outline-story augmented sequences; in generation, the story is sequentially decoded following the same format of augmented sequences. 

Since event sequences are inherently modeled as part of ``the lanague'', FIST model endows the ability to generate succeeding event sequences. During the generation stage, users reserve all flexibility to modify, extend, and manipulate those events right before generating succeeding paragraphs. For instance, users can replace a generated event sequence starting from ``$<\!\!|\text{startofcond}|\!\!>$'' to ``$<\!\!|\text{startofcond}|\!\!>$'' in place by another user written event sequence. FIST model may also wrap up a story indefinitely with less or more paragraphs than the provided length of outline, {\it i.e.}, the number of events. We will demonstrate through experiments the quality and characteristics of model generated outline events. It is all up to an user's practical needs to decide the level of human supervision. Overall, FIST model keeps the highest level of generation controllability and flexibility to fulfill the promise of fine-grained controllable generation.

\section{Related Work}
\subsection{Controllable Text Generation}
A number of previous studies of controllable text generation have focused on certain global aspects of the text. The most common aspects are sentiment and topic \cite{shen2017style,zhao2018adversarially,hu2017toward,fang2019implicit,dathathri2019plug,keskar2019ctrl}. Also, researchers attempt fine-grained control with plots, plans or the so-called storylines \cite{peng2018towards,yao2019plan}, leading to a wide usage and benchmark on 5-lines story dataset $\mathtt{ROCStories}$ \cite{mostafazadeh2016corpus} and similar datasets with relatively short text. Later on, long story generation stands at the frontier \cite{fan2018hierarchical,fan2019strategies} of conditional and controllable generation. The task named as ``neural story generation'' is still under development and relatively under-explored so far.

So far, controllable generation with increasingly longer text and finer attributes has not been studied very well. This paper proposes the ``outline to story'' task to instantiate such research interest with controlled conditions as evolving short text. Recently, a concurrent work \cite{rashkin2020plotmachines} proposes a similar task and a dedicated architecture with memory mechanisms integrated with pre-trained language models. This paper will benchmark both models and promote the application from outline to open-domain long stories.

\subsection{Transfer Learning on Language Models}
Language modeling \cite{bengio2003neural} have played an important role in natural language processing and understanding, especially when used as a pre-training technique. Word embeddings \cite{mikolov2013distributed} and contextual word vectors \cite{mccann2017learned,howard2018universal,peters2018deep} are some early but significant products. Recent works of large Transformer architectures \cite{vaswani2017attention,radford2019language,devlin2018bert} have leveraged the power of both big models and big data to further improve language representation. A number of works also study different pre-training model bases \cite{song2019mass,dong2019unified,keskar2019ctrl,lample2019cross,zellers2019defending}. In terms of generation tasks, the GPT-2 models attracts huge attention due to its dedicated design for unconditional language generation. Researchers \cite{see2019massively,mao2019improving,ziegler2019encoder} have studied transfer learning on GPT-2 models to solve conditional generation tasks. For instance, \cite{ziegler2019encoder} impressively introduces modification to self-attention architecture for adapting a pre-trained model to arbitrary conditional input that goes beyond text. 
However, we note that how to leverage pre-trained language models for fine-grained controllable generation of long text is still an open problem. 

\section{Experiments and Discussions}
\subsection{Experimental Settings}
We first evaluate various models' story generation capability; then evaluate and emphasize controllability and flexibility in generation, a.k.a. production or inference stage.

We conduct experiments on $\mathtt{WritingPrompts}$ and $\mathtt{WikiPlots}$ as introduced in ``the Task and Data'' section, which meet our target of open-domain long text corpora with paired outlines.

We notice that our model generating story from outline doesn't necessarily need prompt as part of input. However, since conventional story generation do uses prompt as input and the concurrent work \cite{rashkin2020plotmachines} also uses prompt as input to form outlines, we always conduct experiments using prompt as an useful addition to perform global control for a fair comparison between models.

We use the smallest public version GPT-2 as a pre-trained model base, which is a large auto-regressive transformer based language model trained on 40 GB of non-Wikipedia text \cite{radford2019language}. It is, by our knowledge, the most widely used model base in relevant literature. Note that there are many other pre-trained language models that may be larger and more powerful than our used one \cite{song2019mass,dong2019unified,keskar2019ctrl,lample2019cross,zellers2019defending}. Our argument is that, the purpose of our experiment is not to compare powers of different pre-trained model bases. It's orthogonal to this work to investigate different pre-trained model bases.  

% CTRL， GROVER 实验部分应该是比较不同的控制方法，而不是基于不同的language modeling base来继续fine tune, 因此不用做这两个实验

\subsubsection{Benchmark Models}
We compare models with designated purposes. By comparing with a specialized-architecture task-specific story generation model \cite{fan2018hierarchical}, we evaluate the model's in-domain generation performances. The fusion model in \cite{fan2018hierarchical} takes a convolutional seq2seq model structure with a fusion training mechanism. 

By comparing with a concurrent work \cite{rashkin2020plotmachines}, we compare FIST with PlotMachines, a dedicated architecture using memory mechanisms for outline-conditioned story generation. Note that the benchmarked PlotMachines use the same GPT2 model base to construct its architecture.

By comparing with state-of-the-art transfer learning method based on GPT-2 models \cite{ziegler2019encoder}, we evaluate different ways to absorb input for conditional generation. The encoder-agnostic model adaptation in \cite{ziegler2019encoder} has certain advantage of absorbing arbitrary conditional input other than text. Its key architecture is the pseudo self attention (PSA), which introduces new projection matrices to absorb input embeddings to the self-attention framework. 

By comparing different inputs, we perform ablation study on input side to evaluate effect of outlines in conditional generation. One input source is only the prompt, whereas the other is ``prompt + outline'' connected with necessary delimiters.

By comparing with another way to build augmented sequence for fine-tuning, we perform ablation study on the FIST model side. The alternative way is the prepending manner shown in Table~\ref{table:augmented_seq}, while our FIST advocates to join cascaded events and paragraphs in an interleaving manner.

\begin{table*}[t]
\begin{centering}
\begin{adjustbox}{scale={0.92}{0.92},center}
\begin{tabular}{c|c|c|c|c|c|c|c|c|c|c|c|c}
\toprule
\multirow{2}{*}{Methods} & \multicolumn{2}{c|}{Perplexity\! $\downarrow$} & \multirow{2}{*}{BLEU-4\! $\uparrow$} & \multicolumn{3}{c|}{ROUGE-1\! $\uparrow$} & \multicolumn{3}{c|}{ROUGE-2\! $\uparrow$} & \multicolumn{3}{c}{ROUGE-L\! $\uparrow$} \tabularnewline
\cline{2-3}\cline{5-13}
 & Word & BPE & & F1 & P & R & F1 & P & R & F1 & P & R \tabularnewline
\toprule
\multicolumn{13}{c}{Dataset: $\mathtt{WritingPrompts}$}\tabularnewline
\toprule
Fusion (only prompt) & 36.0 & - & 0.372 & 0.223 & \textbf{0.386} & 0.157 & 0.038 & 0.074 & 0.026 & 0.206 & \textbf{0.358} & 0.145 \tabularnewline
Fusion (prompt + outline) & 33.3 & - & 0.375 & 0.268 & \textbf{0.455} & 0.19 & 0.063 & \textbf{0.119} & 0.043 & 0.25 & \textbf{0.424} & 0.177 \tabularnewline
\midrule
PSA (only prompt) & 31.6 & 21.3 & \textbf{0.387} & 0.265 & 0.316 & 0.228 & 0.047 & 0.054 & 0.041 & 0.248 & 0.296 & 0.213 \tabularnewline
PSA (prompt + outline) & 30.9 & 20.9 & \textbf{0.389} & 0.265 & 0.327 & 0.223 & 0.048 & 0.056 & 0.042 & 0.249 & 0.307 & 0.209 \tabularnewline
\midrule
PoltMachines & - & - & - & - & - & \textbf{0.311} & - & - & 0.067 & - & - & 0.261 \tabularnewline
\midrule
FIST (only prompt) & 30.2 & 25.9 & 0.356 & 0.181 & 0.339 & 0.123 & 0.023 & 0.046 & 0.015 & 0.17 & 0.321 & 0.116 \tabularnewline
FIST (both, prepend) & \textbf{18.9} & \textbf{16.6} & 0.382 & \textbf{0.299} & 0.324 & 0.277 & \textbf{0.069} & 0.07 & \textbf{0.069} & \textbf{0.283} & 0.308 & \textbf{0.262} \tabularnewline
FIST & \textbf{20.3} & \textbf{17.6} & 0.377 & \textbf{0.294} & 0.347 & 0.255 & \textbf{0.07} & \textbf{0.078} & 0.063 & \textbf{0.279} & 0.33 & 0.242 \tabularnewline
\bottomrule
\multicolumn{13}{c}{Dataset: $\mathtt{WikiPlots}$}\tabularnewline
\toprule
Fusion (only prompt) & 108.2 & - & 0.333 & 0.185 & 0.185 & 0.185 & 0.026 & 0.027 & 0.025 & 0.15 & 0.149 & 0.151 \tabularnewline
Fusion (prompt + outline) & 79.1 & - & 0.342 & 0.232 & 0.244 & 0.221 & 0.057 & 0.059 & 0.056 & 0.185 & 0.194 & 0.177 \tabularnewline
\midrule
PSA (only prompt) & 79.5 & 47.8 & 0.326 & 0.188 & 0.188 & 0.189 & 0.026 & 0.025 & 0.027 & 0.172 & 0.171 & 0.173 \tabularnewline
PSA (prompt + outline) & 79.2 & 47.7 & \textbf{0.344} & 0.185 & 0.186 & 0.185 & 0.024 & 0.023 & 0.026 & 0.167 & 0.168 & 0.167 \tabularnewline
\midrule
PoltMachines & - & - & - & - & - & 0.228 & - & - & 0.065 & - & - & 0.175 \tabularnewline
\midrule
FIST (only prompt) & 38.9 & 26.5 & 0.265 & 0.166 & 0.253 & 0.124 & 0.018 & 0.032 & 0.013 & 0.15 & 0.231 & 0.111 \tabularnewline
FIST (both, prepend) & \textbf{26.0} & \textbf{18.5} & \textbf{0.346} & \textbf{0.281} & \textbf{0.284} & \textbf{0.279} & \textbf{0.083} & \textbf{0.081} & \textbf{0.086} & \textbf{0.254} & \textbf{0.257} & \textbf{0.252} \tabularnewline
FIST & \textbf{26.7} & \textbf{18.9} & 0.333 & \textbf{0.275} & \textbf{0.289} & \textbf{0.262} & \textbf{0.081} & \textbf{0.084} & \textbf{0.078} & \textbf{0.248} & \textbf{0.261} & \textbf{0.237} \tabularnewline
\bottomrule
\end{tabular}
\end{adjustbox}
\end{centering}
\caption{Automatic metrics for conditional story generation evaluated on two datasets. }
\label{table:automatic_metrics}
\end{table*}

\subsubsection{Implementation Details}
We implement FIST using the ``Huggingface Transformers'' library in Pytorch \cite{Wolf2019HuggingFacesTS}. Special tokens are conveniently added to the original 50K BPE GPT-2 vocabulary. All GPT-2 model settings remain the same. Other models are re-trained and re-evaluated with the same datasets using their own implementation repositories if needed. In evaluation, we generate stories using the top-k top-p random sampling scheme \cite{holtzman2019curious,keskar2019ctrl} with $k=100$ and $p=0.9$. Temperature smoothing technique is also applied with $T=0.9$. 

Considering two relatively large test datasets, we only decode one story per test input. When testing an outline-story pair using FIST, we replace the model generated outline event in place by the corresponding automatically extracted outline event from the test story.

\begin{table*}[t]
\begin{centering}
\begin{adjustbox}{scale={0.97}{0.97},center}
\begin{tabular}{c|c|c|c|c|c|c|c|c|c|c|c|c|c|c}
\toprule
\multirow{2}{*}{Decoding} & \multirow{2}{*}{\!\!Story\!\!} & \multirow{2}{*}{\!\!Event\!\!} & \multirow{2}{*}{\!\!Phrase\!\!} & \multicolumn{2}{c|}{BLEU-4\! $\uparrow$} & \multicolumn{3}{c|}{ROUGE-1\! $\uparrow$} & \multicolumn{3}{c|}{ROUGE-2\! $\uparrow$} & \multicolumn{3}{c}{ROUGE-L\! $\uparrow$} \tabularnewline
\cline{5-15}
\!\!scenario & para. & \!\!phrase\!\! & len. & story & event & F1 & P & R & F1 & P & R & F1 & P & R \tabularnewline
\toprule
\multicolumn{15}{c}{Dataset: $\mathtt{WritingPrompts}$}\tabularnewline
\toprule
FIST-$a$ & 2.9 & 2.4 & 3.0 & 0.377 & 0.13 & 0.22 & 0.333 & 0.2 & 0.044 & 0.061 & 0.044 & 0.184 & 0.318 & 0.19 \tabularnewline
\midrule
FIST-$r$ & 2.9 & 2.4 & 3.0 & 0.371 & 0.12 & 0.21 & 0.323 & 0.191 & 0.039 & 0.055 & 0.038 & 0.173 & 0.308 & 0.181 \tabularnewline
\midrule
FIST-$f$ & 3.0 & 2.4 & 2.9 & 0.358 & 0.04 & 0.21 & 0.328 & 0.187 & 0.038 & 0.055 & 0.037 & 0.172 & 0.313 & 0.178 \tabularnewline
\bottomrule
\multicolumn{15}{c}{Dataset: $\mathtt{WikiPlots}$}\tabularnewline
\toprule
FIST-$a$ & 2.9 & 2.6 & 3.3 & 0.333 & 0.21 & 0.239 & 0.263 & 0.246 & 0.058 & 0.062 & 0.065 & 0.203 & 0.24 & 0.225 \tabularnewline
\midrule
FIST-$r$ & 2.9 & 2.6 & 3.3 & 0.325 & 0.20 & 0.229 & 0.254 & 0.239 & 0.053 & 0.057 & 0.061 & 0.191 & 0.232 & 0.218 \tabularnewline
\midrule
FIST-$f$ & 2.9 & 2.5 & 3.2 & 0.321 & 0.06 & 0.229 & 0.257 & 0.234 & 0.053 & 0.058 & 0.059 & 0.192 & 0.234 & 0.213 \tabularnewline
\bottomrule
\end{tabular}
\end{adjustbox}
\end{centering}
\caption{Automatic metrics for event analysis.}
\label{table:event_generation}
\end{table*}

\subsection{Automatic and Qualitative Evaluation}
\subsubsection{Automatic Metrics} We basically evaluate the following automatic metrics towards target stories:
\begin{itemize}
	\item \textbf{Perplexity (PPL)} is used to evaluate language models and often regarded as a proxy for generation quality. All models based on GPT-2 use BPE tokenization scheme, where PPL values are not directly comparable with some previous models such as \cite{fan2018hierarchical}, where PPLs are computed at natural word level. Similarly to \cite{see2019massively}, we additionally compute word-level perplexity of GPT-2 models to enable the comparison with previous models. That is, we normalize the total negative log probability of the target text by the number of word level tokens. To ensure the fairness, when we evaluate perplexity on augmented sequences using FIST, we only count over story tokens, but not tokens of the event sequences, since other models only count over story tokens.
	\item \textbf{BLEU and ROUGE scores} are computed as n-gram overlap of generated stories versus target stories. We compute 4-gram BLEU using the ``NLTK'' library and ROUGE scores (ROUGE-1, ROUGE-2, and ROUGE-L) using \cite{lin2002manual}. The ROUGE score includes both precision (P), recall (R), and F1, with ROUGE precision having similar interpretation as BLEU.
\end{itemize}

%\vspace{-2.0mm}
\subsubsection{Automatic Evaluation Results}
The automatic evaluation results are presented in Table~\ref{table:automatic_metrics}. Overall, the proposed FIST with full input source (prompt and outline) generally achieves better results, with lower PPL, higher BLEU and higher ROUGE scores, demonstrating state-of-the-art conditional story generation performance. 

Methods based on pre-trained models, i.e. PSA / PlotMachines / FIST, show better overall performance with relatively less task specific efforts than Fusion models. This demonstrates the power of large-scale pre-training. Fusion models still show relatively high precision scores in $\mathtt{WritingPrompts}$, due to its dedicated design on this dataset.

FIST achieves comparable or even better results over PlotMachines, presenting a much simpler approach based on solely language modeling rather than dedicated architecture. The comparison so far on automatic metrics poses an interesting and open-ended question that, does dedicated architecture really benefit as expected?  

With full input sources, the FIST model generally outperforms PSA, showing that simple language modeling on naturally co-occurred text conditions could grasp conditional dependence to some considerable extent. In other words, architecture modification is not always necessary with text conditions. With only prompt, FIST cannot beat PSA, showing that co-occurred structures labeled by special tokens could work poorly with insufficient information and structure exposure. Therefore, the FIST method may prefer scenarios that text conditions are strong and fruitful, for instance, longer and finer input condition.

The performance gap caused by two different input sources vary between different models: both Fusion models and PSA have limited performance improvement when provided with full inputs, while the FIST model is observed with a relatively larger performance gap. It is reasonable since the FIST model works directly and universally on augmented sequences, which is more sensitive and easier to reflect the enhancement of input information; while others ``separate'' input and story, and are less sensitive to the input.

The FIST model with interleaving joining mechanism shows a little performance drop comparing with the prepending manner, due to the fact that the latter exposes more information along the decoding. We will see that the performance drop is a quite acceptable sacrifice, as FIST model with interleaving joining mechanism achieves great decoding flexibility and controllability.

\subsubsection{Qualitative Evaluation}
We also present some generation examples on the test datasets in Tables~\ref{table:example_a_wp} and \ref{table:example_a_wi} in the Appendix. Stories are seen as semantically and grammatically sound, moreover, highly conditioned on and consistent with given outlines. Due to the pairing nature between extracted outlines and stories, the model presents obvious grasp on at least a lexical level. Tables~\ref{table:example_r_wp}, \ref{table:example_f_wp}, \ref{table:example_r_wi}, \ref{table:example_f_wi} are also good examples. A large scale human evaluation is underway, which is quite expensive due to text length and scale.

\subsection{Analysis on Automatically Generated Events}
Since event sequences are inherently modeled as a part of ``the language'', FIST model has its own ability to generate succeeding event sequences, which future story paragraphs and events are conditioned on. We perform event analysis under various decoding scenarios that involve different levels of event supervision and utilization. 
\begin{itemize}
	\item \textbf{FIST-$a$:} exactly as the test setting in previous evaluation section, the prompt and all the automatically extracted outline events are given following the chronological order.
	\item \textbf{FIST-$r$:} with the prompt and the first event sequence given, the following test events are also given, but randomly shuffled in order.
	\item \textbf{FIST-$f$:} with only the prompt and the first event sequence given, the model will depend on its own generated event sequence starting from the second paragraph till the end.
\end{itemize}

We compare automatic metrics of generated stories in different decoding scenarios versus target stories in Table~\ref{table:event_generation}. Metrics include certain statistics in Table~\ref{table:datasets} and Table~\ref{table:automatic_metrics}, with an extra BLEU-4 score between model generated and target extracted event sequences. 

We observe that tatistics about paragraphs and phrases are quite uniform over different decoding scenarios, implying that the model could generate events in the correct format. Since the GPT-2 model allows a limited context window of at most 1024 tokens, generated stories with different decoding scenarios all consist of roughly 3 paragraphs. From FIST-$a$ to FIST-$r$ and finally FIST-$f$, less coincidence between generated and target event sequences are observed, along with steady but minor metrics drop. We conclude that in terms of capturing content and semantics flowing in target stories, events automatically extracted from target stories have certain advantage, while model generated events show close and considerable semantic consistence. Note that the comparison may be biased since the limited context window may lead to a less metric gap.

In the Appendix, generation examples from the two new introduced decoding scenarios are presented in Tables~\ref{table:example_r_wp}, \ref{table:example_f_wp}, \ref{table:example_r_wi}, \ref{table:example_f_wi}. We colorize target extracted events into red and model generated events into blue. Reasonable quality drop is observed from FIST-$a$ to FIST-$r$ and finally FIST-$f$. 

To emphasize, with the ability of feeding in events in chronological order and generating events automatically, the FIST model has its unique advantage in flexibility and controllability than others, including the PlotMachines.

\section{Conclusion}
In this paper, we propose and create dataset for the task of ``outline to story'' to instantiate research interest of fine-grained controllable generation of open-domain long text with short-text controlled conditions. By leveraging pre-trained language models, our method FIST represents an extremely simple yet strong baseline for the task. More efforts towards the challenging case of open-domain long text should be invested in future works.

% \section*{Acknowledgements}
% We acknowledge our colleagues for helpful discussions and the reviewers ahead for their comments. The authors appreciate peer researchers for sharing their source codes publicly online, which accelerate our experimentation.

% \bibliographystyle{aaai21}
\bibliography{refs}

\clearpage{} 

\onecolumn
\appendix

\vspace*{\fill}
\begin{center}\LARGE{\textbf{Appendix}}\end{center}
\vspace*{\fill}

%\section*{Generation Examples}
%\footnotesize{} \scriptsize{} 

\begin{table*}[htbp] \scriptsize{}
\begin{tcolorbox}
\hspace{-7mm}
\begin{tabular}{l p{0.95\linewidth}}
{\bf \textit{Target}} & ----------------------------------------------------------------------------------------------------------------------------------------------- \\

{\bf Prompt:} & You discover a grand hall filled with legendary weapons like Mjonir and Excalibur. Each generation or so, warriors come to the hall to inherit a weapon that they are worthy enough to wield. Across the hall you see a forgotten weapon that's been collecting dust. You hear it call to you. \\

{\bf Event:} & $<\!\!|\text{startofcond}|\!\!>$ \textcolor{red}{searing pain manifested} $<\!\!|\text{sepofcond}|\!\!>$ \textcolor{red}{dusty forgotten hall} $<\!\!|\text{sepofcond}|\!\!>$ \textcolor{red}{greetings child} $<\!\!|\text{endofcond}|\!\!>$ \\

{\bf Paragraph:} &  A \textcolor{red}{searing pain manifested} in the back of Renards skull, jarring him awake. He found himself lying in a \textcolor{red}{dusty forgotten hall} of stone. "How did I get here?" He asked himself. But no memeories were forthcoming.
 "\textcolor{red}{Greetings child}." The voice sent a chill down his spine and somehow relieved the ache in his head.
 "Whos there!?" He asked loudly into the dimness.
 "Come find me and all shall be clear." \\

{\bf Event:} & $<\!\!|\text{startofcond}|\!\!>$ \textcolor{red}{seemingly holy weapons} $<\!\!|\text{sepofcond}|\!\!>$ \textcolor{red}{renard stood} $<\!\!|\text{endofcond}|\!\!>$ \\

{\bf Paragraph:} & \textcolor{red}{Renard stood}, and against his better will began following what seemed to be the source of the voice. As he walked, his eyes fell upon many ancient and \textcolor{red}{seemingly holy weapons}. Each he passed more obscure than the last, until he could no longer name them. "Closer my child, I am almost in reach." \\

{\bf Event:} & $<\!\!|\text{startofcond}|\!\!>$ \textcolor{red}{one solid blast} $<\!\!|\text{sepofcond}|\!\!>$ \textcolor{red}{mournful baritone echoed} $<\!\!|\text{sepofcond}|\!\!>$ \textcolor{red}{clothes burned away} $<\!\!|\text{endofcond}|\!\!>$ \\

{\bf Paragraph:} & Far and away from the other weapons was a black horn. Simple and eerie. Without any confirmation he knew what must be done. He brought it to his lips and blew, \textcolor{red}{one solid blast}. Its \textcolor{red}{mournful baritone echoed} and somehow agitated the other weapons. Renard collapsed to the ground shrieking as he felt his body covered in invisible flames. His \textcolor{red}{clothes burned away} but no matter how much pain he felt the flames did not damage his body. \\

{\bf Event:} & $<\!\!|\text{startofcond}|\!\!>$ \textcolor{red}{new existence brought forth} $<\!\!|\text{sepofcond}|\!\!>$ \textcolor{red}{go forth} $<\!\!|\text{sepofcond}|\!\!>$ \textcolor{red}{salt} $<\!\!|\text{endofcond}|\!\!>$ \\

{\bf Paragraph:} & Hours, days, weeks, time had fallen to the wayside in the wake of this endless torment. Until suddenly it stopped. He rose, naked in the hall of legend, a \textcolor{red}{new existence brought forth} in pain. His name forgotten only the horn remained. "The age of man has reached its pinnacle." The horn whispered. "\textcolor{red}{Go forth} and bring cleansing. My King of \textcolor{red}{Salt}." \\

{\bf Event:} & $<\!\!|\text{startofcond}|\!\!>$ \textcolor{red}{saw several others storm} $<\!\!|\text{sepofcond}|\!\!>$ \textcolor{red}{stunned londoners stared} $<\!\!|\text{sepofcond}|\!\!>$ \textcolor{red}{first comes destruction} $<\!\!|\text{endofcond}|\!\!>$ \\

{\bf Paragraph:} & Clothing appeared from nonexistence to cover his form. With a mere flick of his wrist the wall broke and he found himself in London. "\textcolor{red}{First comes destruction}, I create nothingness." \textcolor{red}{stunned Londoners stared} at him. Woth a broad wave of his hand they disintegrated into piles of salt. He rose to the sky on blackened wings. Below he \textcolor{red}{saw several others storm} out of the hall, holding the feeble weapons built in a bygone era to destroy him. \\

{\bf Event:} & $<\!\!|\text{startofcond}|\!\!>$ \textcolor{red}{human disease} $<\!\!|\text{sepofcond}|\!\!>$ \textcolor{red}{voice boomed} $<\!\!|\text{endofcond}|\!\!>$  \\

{\bf Paragraph:} & His \textcolor{red}{voice boomed} over the city like the crashing of thunder. "Hail the arisen king for the end of days is not as god prophesized. And cower for the land will be cleansed of the \textcolor{red}{human disease}." 
\\
{\bf \textit{Generation}} & ----------------------------------------------------------------------------------------------------------------------------------------------- \\

{\bf Prompt:} & You discover a grand hall filled with legendary weapons like Mjonir and Excalibur. Each generation or so, warriors come to the hall to inherit a weapon that they are worthy enough to wield. Across the hall you see a forgotten weapon that's been collecting dust. You hear it call to you. \\

{\bf Event:} & $<\!\!|\text{startofcond}|\!\!>$ \textcolor{red}{searing pain manifested} $<\!\!|\text{sepofcond}|\!\!>$ \textcolor{red}{dusty forgotten hall} $<\!\!|\text{sepofcond}|\!\!>$ \textcolor{red}{greetings child} $<\!\!|\text{endofcond}|\!\!>$ \\

{\bf Paragraph:} & A \textcolor{red}{searing pain manifested} from the mauve handle of my mighty sword and as it pierced the air, I knew I would be rewarded for my courage. I had come here at the wrong time. I looked around the \textcolor{red}{dusty forgotten hall} for any signs of the many who had come before me. I would have no fear. I would have a rest. A rest for my soul and the rest for my friends and family. 
 "Hello \textcolor{red}{child}. You have reached the Hall of \textcolor{red}{Greetings}." \\

{\bf Event:} & $<\!\!|\text{startofcond}|\!\!>$ \textcolor{red}{seemingly holy weapons} $<\!\!|\text{sepofcond}|\!\!>$ renard stood $<\!\!|\text{endofcond}|\!\!>$ \\

{\bf Paragraph:} & I looked around the hall. The hall was empty. All the hall had seemed to be covered in \textcolor{red}{seemingly holy weapons}. But what could I get for my sword?
 "I know you have come here once before, but you seem to have forgotten. What is this?" I said to the floor. 
 "You are no match for our halls. Please, come forth and slay my foes." 
 ...... 
 "Sirard! Are you alright?"ard said to the floor. \\

{\bf Event:} & $<\!\!|\text{startofcond}|\!\!>$ \textcolor{red}{one solid blast} $<\!\!|\text{sepofcond}|\!\!>$ \textcolor{red}{mournful baritone echoed} $<\!\!|\text{sepofcond}|\!\!>$ \textcolor{red}{clothes burned away} $<\!\!|\text{endofcond}|\!\!>$ \\

{\bf Paragraph:} & I raised my sword high. The cheers of the crowd \textcolor{red}{echoed}. The hall grew in size. I raised it up to my head. I could see the blood of my men. But what could I do with that? I raised it again. I could see that they had not lost a single life. The wound on the chest had been severed. The \textcolor{red}{clothing burned away}. I raised the sword to my head. 
 "You have defeated my foes," the \textcolor{red}{mournful baritone echoed} through the hall. "There is only \textcolor{red}{one solid blast} of energy from your enemies that will awaken them." \\

{\bf Event:} & $<\!\!|\text{startofcond}|\!\!>$ \textcolor{red}{new existence brought forth} $<\!\!|\text{sepofcond}|\!\!>$ \textcolor{red}{go forth} $<\!\!|\text{sepofcond}|\!\!>$ salt $<\!\!|\text{endofcond}|\!\!>$ \\

{\bf Paragraph:} & But what happened next? Was I really the same person I had been before? My \textcolor{red}{new existence brought forth} the death of my people. The only thing that could have saved the humans I believed in is the secret to immortality. 
 "\textcolor{red}{Go forth}!" I shouted. I charged my sword. The noise was almost deafening. The battle was over. 
 "My friends," I said to the ground. "My friends are gone. My friend, who is this?" 
 "I am the last of my kind. You have defeated my enemy." 
 "Good. But how? I'm just a guard that looks after my people. You should have heard me say that. But I knew. I knew. I knew that you were the last hope I had. And now here I stand." \\

{\bf Event:} & $<\!\!|\text{startofcond}|\!\!>$ \textcolor{red}{saw several others storm} $<\!\!|\text{sepofcond}|\!\!>$ \textcolor{red}{stunned londoners stared} $<\!\!|\text{sepofcond}|\!\!>$ \textcolor{red}{first comes destruction} $<\!\!|\text{endofcond}|\!\!>$ \\

{\bf Paragraph:} & I looked at my enemy. He stood over me. I \textcolor{red}{saw several others storm} over me. I did not remember much. I tried to remember myself. How I used to kill them. Or how I had saved them. I did not remember how to fight. I was just a guard trying to protect my people. But I was suddenly aware of the fear I had. I looked up at my enemy. He was now wearing a bloodied, battered looking uniform. I did not remember how I knew to kill one of them. 
 "This is your chance. The \textcolor{red}{first comes destruction}. The next is vengeance." 
 The \textcolor{red}{stunned Londoners stared} at me. They looked at me with confusion. I nodded. "I will be one of them." \\

{\bf Event:} &  $<\!\!|\text{startofcond}|\!\!>$ \textcolor{red}{human disease} $<\!\!|\text{sepofcond}|\!\!>$ \textcolor{red}{voice boomed} $<\!\!|\text{endofcond}|\!\!>$ \\

{\bf Paragraph:} &  I didn't understand what they were saying. How could I fight off a man in black? All I could hear was the clatter of hooves on stone. 
 A \textcolor{red}{voice boomed} from the sky. "I am the last of my kind. Welcome to the \textcolor{red}{Human Disease}."
\end{tabular}
\end{tcolorbox}
\caption{Generation examples on test dataset of $\mathtt{WritingPrompts}$ with FIST-$a$ decoding.}
\label{table:example_a_wp}
\end{table*}

\clearpage{} 

\begin{table*}[htbp] \footnotesize{} 
\begin{tcolorbox}
\hspace{-7mm}
\begin{tabular}{l p{0.92\linewidth}}

{\bf \textit{Target}} & --------------------------------------------------------------------------------------------------------------------------------- \\

{\bf Prompt:} & You discover a grand hall filled with legendary weapons like Mjonir and Excalibur. Each generation or so, warriors come to the hall to inherit a weapon that they are worthy enough to wield. Across the hall you see a forgotten weapon that's been collecting dust. You hear it call to you. \\

{\bf Event:} & $<\!\!|\text{startofcond}|\!\!>$ \textcolor{red}{searing pain manifested} $<\!\!|\text{sepofcond}|\!\!>$ \textcolor{red}{dusty forgotten hall} $<\!\!|\text{sepofcond}|\!\!>$ \textcolor{red}{greetings child} $<\!\!|\text{endofcond}|\!\!>$ \\

{\bf Paragraph:} &  A \textcolor{red}{searing pain manifested} in the back of Renards skull, jarring him awake. He found himself lying in a \textcolor{red}{dusty forgotten hall} of stone. "How did I get here?" He asked himself. But no memeories were forthcoming.
 "\textcolor{red}{Greetings child}." The voice sent a chill down his spine and somehow relieved the ache in his head.
 "Whos there!?" He asked loudly into the dimness.
 "Come find me and all shall be clear." \\

{\bf Event:} & $<\!\!|\text{startofcond}|\!\!>$ \textcolor{red}{seemingly holy weapons} $<\!\!|\text{sepofcond}|\!\!>$ \textcolor{red}{renard stood} $<\!\!|\text{endofcond}|\!\!>$ \\

{\bf Paragraph:} & \textcolor{red}{Renard stood}, and against his better will began following what seemed to be the source of the voice. As he walked, his eyes fell upon many ancient and \textcolor{red}{seemingly holy weapons}. Each he passed more obscure than the last, until he could no longer name them. "Closer my child, I am almost in reach." \\

{\bf Event:} & $<\!\!|\text{startofcond}|\!\!>$ \textcolor{red}{one solid blast} $<\!\!|\text{sepofcond}|\!\!>$ \textcolor{red}{mournful baritone echoed} $<\!\!|\text{sepofcond}|\!\!>$ \textcolor{red}{clothes burned away} $<\!\!|\text{endofcond}|\!\!>$ \\

{\bf Paragraph:} & Far and away from the other weapons was a black horn. Simple and eerie. Without any confirmation he knew what must be done. He brought it to his lips and blew, \textcolor{red}{one solid blast}. Its \textcolor{red}{mournful baritone echoed} and somehow agitated the other weapons. Renard collapsed to the ground shrieking as he felt his body covered in invisible flames. His \textcolor{red}{clothes burned away} but no matter how much pain he felt the flames did not damage his body. \\

{\bf Event:} & $<\!\!|\text{startofcond}|\!\!>$ \textcolor{red}{new existence brought forth} $<\!\!|\text{sepofcond}|\!\!>$ \textcolor{red}{go forth} $<\!\!|\text{sepofcond}|\!\!>$ \textcolor{red}{salt} $<\!\!|\text{endofcond}|\!\!>$ \\

{\bf Paragraph:} & Hours, days, weeks, time had fallen to the wayside in the wake of this endless torment. Until suddenly it stopped. He rose, naked in the hall of legend, a \textcolor{red}{new existence brought forth} in pain. His name forgotten only the horn remained. "The age of man has reached its pinnacle." The horn whispered. "\textcolor{red}{Go forth} and bring cleansing. My King of \textcolor{red}{Salt}." \\

{\bf Event:} & $<\!\!|\text{startofcond}|\!\!>$ \textcolor{red}{saw several others storm} $<\!\!|\text{sepofcond}|\!\!>$ \textcolor{red}{stunned londoners stared} $<\!\!|\text{sepofcond}|\!\!>$ \textcolor{red}{first comes destruction} $<\!\!|\text{endofcond}|\!\!>$ \\

{\bf Paragraph:} & Clothing appeared from nonexistence to cover his form. With a mere flick of his wrist the wall broke and he found himself in London. "\textcolor{red}{First comes destruction}, I create nothingness." \textcolor{red}{stunned Londoners stared} at him. Woth a broad wave of his hand they disintegrated into piles of salt. He rose to the sky on blackened wings. Below he \textcolor{red}{saw several others storm} out of the hall, holding the feeble weapons built in a bygone era to destroy him. \\

{\bf Event:} & $<\!\!|\text{startofcond}|\!\!>$ \textcolor{red}{human disease} $<\!\!|\text{sepofcond}|\!\!>$ \textcolor{red}{voice boomed} $<\!\!|\text{endofcond}|\!\!>$  \\

{\bf Paragraph:} & His \textcolor{red}{voice boomed} over the city like the crashing of thunder. "Hail the arisen king for the end of days is not as god prophesized. And cower for the land will be cleansed of the \textcolor{red}{human disease}." \\
\\
{\bf \textit{Generation}} & --------------------------------------------------------------------------------------------------------------------------------- \\

{\bf Prompt:} & You discover a grand hall filled with legendary weapons like Mjonir and Excalibur. Each generation or so, warriors come to the hall to inherit a weapon that they are worthy enough to wield. Across the hall you see a forgotten weapon that's been collecting dust. You hear it call to you. \\

{\bf Event:} & $<\!\!|\text{startofcond}|\!\!>$ \textcolor{red}{searing pain manifested} $<\!\!|\text{sepofcond}|\!\!>$ \textcolor{red}{dusty forgotten hall} $<\!\!|\text{sepofcond}|\!\!>$ \textcolor{red}{greetings child} $<\!\!|\text{endofcond}|\!\!>$ \\

{\bf Paragraph:} & A \textcolor{red}{searing pain manifested} from the mauve handle of my mighty sword and as it pierced the air, I knew I would be rewarded for my courage. I had come here at the wrong time. I looked around the \textcolor{red}{dusty forgotten hall} for any signs of the many who had come before me. I would have no fear. I would have a rest. A rest for my soul and the rest for my friends and family. 
 "Hello \textcolor{red}{child}. You have reached the Hall of \textcolor{red}{Greetings}." \\

{\bf Event:} & $<\!\!|\text{startofcond}|\!\!>$ \textcolor{red}{new existence brought forth} $<\!\!|\text{sepofcond}|\!\!>$ \textcolor{red}{go forth} $<\!\!|\text{sepofcond}|\!\!>$ \textcolor{red}{salt} $<\!\!|\text{endofcond}|\!\!>$ \\

{\bf Paragraph:} & I stared at my mother, looking over her shoulder at the men and women I had come to serve. They had come from many places. They had brought with them knowledge, from each great nation to its depths and it was all worth it. They were here to bring me freedom, and the next time, they would make it into the Hall of \textcolor{red}{Salt}... This time, however, my \textcolor{red}{new existence brought forth} a sadness. I wasn't ready. I was ready to take a break. I didn't want to \textcolor{red}{go forth}... \\

{\bf Event:} & $<\!\!|\text{startofcond}|\!\!>$ \textcolor{red}{saw several others storm} $<\!\!|\text{sepofcond}|\!\!>$ \textcolor{red}{stunned londoners stared} $<\!\!|\text{sepofcond}|\!\!>$ \textcolor{red}{first comes destruction} $<\!\!|\text{endofcond}|\!\!>$ \\

{\bf Paragraph:} & Suddenly, I \textcolor{red}{saw several others storm} from my little house, with weapons of all types and from all different stories. The \textcolor{red}{first comes destruction}, the second comes joy. I stepped out of my chair to find the first to be wounded and fallen, and the last to be killed by the first. As the horrified and \textcolor{red}{stunned Londoners stared} in shock and fear at the blade of the first, I heard the familiar voice of my mother speak to me. 
 "I'm sorry. I didn't mean to... But this is an honor and our journey isn't over yet." \\

{\bf Event:} & $<\!\!|\text{startofcond}|\!\!>$ \textcolor{red}{human disease} $<\!\!|\text{sepofcond}|\!\!>$ \textcolor{red}{voice boomed} $<\!\!|\text{endofcond}|\!\!>$ \\

{\bf Paragraph:} & My mother screamed and wailed, as the blade glowed from the eyes of the blade. A \textcolor{red}{voice boomed} over the commotion. 
 "This is your one chance. Please help me, I have lost the last of my friends. Let me rest, I want to protect the \textcolor{red}{human disease}." \\
\end{tabular}
\end{tcolorbox}
\caption{Generation examples on test dataset of $\mathtt{WritingPrompts}$ with FIST-$r$ decoding.}
\label{table:example_r_wp}
\end{table*}

\clearpage{} 

\begin{table*}[htbp] \tiny{}
\begin{tcolorbox}
\hspace{-7mm}
\begin{tabular}{l p{0.96\linewidth}}

{\bf \textit{Target}} & --------------------------------------------------------------------------------------------------------------------------------------------------------------------------- \\

{\bf Prompt:} & You discover a grand hall filled with legendary weapons like Mjonir and Excalibur. Each generation or so, warriors come to the hall to inherit a weapon that they are worthy enough to wield. Across the hall you see a forgotten weapon that's been collecting dust. You hear it call to you. \\

{\bf Event:} & $<\!\!|\text{startofcond}|\!\!>$ \textcolor{red}{searing pain manifested} $<\!\!|\text{sepofcond}|\!\!>$ \textcolor{red}{dusty forgotten hall} $<\!\!|\text{sepofcond}|\!\!>$ \textcolor{red}{greetings child} $<\!\!|\text{endofcond}|\!\!>$ \\

{\bf Paragraph:} &  A \textcolor{red}{searing pain manifested} in the back of Renards skull, jarring him awake. He found himself lying in a \textcolor{red}{dusty forgotten hall} of stone. "How did I get here?" He asked himself. But no memeories were forthcoming.
 "\textcolor{red}{Greetings child}." The voice sent a chill down his spine and somehow relieved the ache in his head.
 "Whos there!?" He asked loudly into the dimness.
 "Come find me and all shall be clear." \\

{\bf Event:} & $<\!\!|\text{startofcond}|\!\!>$ \textcolor{red}{seemingly holy weapons} $<\!\!|\text{sepofcond}|\!\!>$ \textcolor{red}{renard stood} $<\!\!|\text{endofcond}|\!\!>$ \\

{\bf Paragraph:} & \textcolor{red}{Renard stood}, and against his better will began following what seemed to be the source of the voice. As he walked, his eyes fell upon many ancient and \textcolor{red}{seemingly holy weapons}. Each he passed more obscure than the last, until he could no longer name them. "Closer my child, I am almost in reach." \\

{\bf Event:} & $<\!\!|\text{startofcond}|\!\!>$ \textcolor{red}{one solid blast} $<\!\!|\text{sepofcond}|\!\!>$ \textcolor{red}{mournful baritone echoed} $<\!\!|\text{sepofcond}|\!\!>$ \textcolor{red}{clothes burned away} $<\!\!|\text{endofcond}|\!\!>$ \\

{\bf Paragraph:} & Far and away from the other weapons was a black horn. Simple and eerie. Without any confirmation he knew what must be done. He brought it to his lips and blew, \textcolor{red}{one solid blast}. Its \textcolor{red}{mournful baritone echoed} and somehow agitated the other weapons. Renard collapsed to the ground shrieking as he felt his body covered in invisible flames. His \textcolor{red}{clothes burned away} but no matter how much pain he felt the flames did not damage his body. \\

{\bf Event:} & $<\!\!|\text{startofcond}|\!\!>$ \textcolor{red}{new existence brought forth} $<\!\!|\text{sepofcond}|\!\!>$ \textcolor{red}{go forth} $<\!\!|\text{sepofcond}|\!\!>$ \textcolor{red}{salt} $<\!\!|\text{endofcond}|\!\!>$ \\

{\bf Paragraph:} & Hours, days, weeks, time had fallen to the wayside in the wake of this endless torment. Until suddenly it stopped. He rose, naked in the hall of legend, a \textcolor{red}{new existence brought forth} in pain. His name forgotten only the horn remained. "The age of man has reached its pinnacle." The horn whispered. "\textcolor{red}{Go forth} and bring cleansing. My King of \textcolor{red}{Salt}." \\

{\bf Event:} & $<\!\!|\text{startofcond}|\!\!>$ \textcolor{red}{saw several others storm} $<\!\!|\text{sepofcond}|\!\!>$ \textcolor{red}{stunned londoners stared} $<\!\!|\text{sepofcond}|\!\!>$ \textcolor{red}{first comes destruction} $<\!\!|\text{endofcond}|\!\!>$ \\

{\bf Paragraph:} & Clothing appeared from nonexistence to cover his form. With a mere flick of his wrist the wall broke and he found himself in London. "\textcolor{red}{First comes destruction}, I create nothingness." \textcolor{red}{stunned Londoners stared} at him. Woth a broad wave of his hand they disintegrated into piles of salt. He rose to the sky on blackened wings. Below he \textcolor{red}{saw several others storm} out of the hall, holding the feeble weapons built in a bygone era to destroy him. \\

{\bf Event:} & $<\!\!|\text{startofcond}|\!\!>$ \textcolor{red}{human disease} $<\!\!|\text{sepofcond}|\!\!>$ \textcolor{red}{voice boomed} $<\!\!|\text{endofcond}|\!\!>$  \\

{\bf Paragraph:} & His \textcolor{red}{voice boomed} over the city like the crashing of thunder. "Hail the arisen king for the end of days is not as god prophesized. And cower for the land will be cleansed of the \textcolor{red}{human disease}."
\\
{\bf \textit{Generation}} & --------------------------------------------------------------------------------------------------------------------------------------------------------------------------- \\

{\bf Prompt:} & You discover a grand hall filled with legendary weapons like Mjonir and Excalibur. Each generation or so, warriors come to the hall to inherit a weapon that they are worthy enough to wield. Across the hall you see a forgotten weapon that's been collecting dust. You hear it call to you. \\

{\bf Event:} & $<\!\!|\text{startofcond}|\!\!>$ \textcolor{red}{searing pain manifested} $<\!\!|\text{sepofcond}|\!\!>$ \textcolor{red}{dusty forgotten hall} $<\!\!|\text{sepofcond}|\!\!>$ \textcolor{red}{greetings child} $<\!\!|\text{endofcond}|\!\!>$ \\

{\bf Paragraph:} & A \textcolor{red}{searing pain manifested} from the mauve handle of my mighty sword and as it pierced the air, I knew I would be rewarded for my courage. I had come here at the wrong time. I looked around the \textcolor{red}{dusty forgotten hall} for any signs of the many who had come before me. I would have no fear. I would have a rest. A rest for my soul and the rest for my friends and family. 
 "Hello \textcolor{red}{child}. You have reached the Hall of \textcolor{red}{Greetings}." \\

{\bf Event:} & $<\!\!|\text{startofcond}|\!\!>$ \textcolor{blue}{small red rock} $<\!\!|\text{sepofcond}|\!\!>$ \textcolor{blue}{lunge closer} $<\!\!|\text{sepofcond}|\!\!>$ \textcolor{blue}{first time} $<\!\!|\text{endofcond}|\!\!>$ \\

{\bf Paragraph:} & There was no response. No movement. No sounds. I took a few steps and prepared myself. I drew my sword and prepared to strike at any who would dare to stand before me. The wall opened up in front of me and I \textcolor{blue}{lunge closer}. It was clear. In a flash of light I could see a \textcolor{blue}{small red rock} with blue flames dancing in the flame. Instinctively I drew the blade and drew my blade. 
 "Why have you come here? Why have you come here." \\

{\bf Event:} & $<\!\!|\text{startofcond}|\!\!>$ \textcolor{blue}{sword seemed} $<\!\!|\text{sepofcond}|\!\!>$ \textcolor{blue}{time} $<\!\!|\text{endofcond}|\!\!>$ \\

{\bf Paragraph:} & I looked up at the flame and saw it was the fires of the damned, and the souls of all who died. I had had enough. I used my \textcolor{blue}{time} to move the sword and with a simple heave I stabbed it into the flame. In the blink of an eye, the \textcolor{blue}{sword seemed} to melt. \\

{\bf Event:} & $<\!\!|\text{startofcond}|\!\!>$ \textcolor{blue}{one last time} $<\!\!|\text{sepofcond}|\!\!>$ \textcolor{blue}{one moment} $<\!\!|\text{endofcond}|\!\!>$ \\

{\bf Paragraph:} & I charged forward, aiming for the beast before me, and \textcolor{blue}{one moment} I was ready. I took a moment to close my eyes and breathe a sigh of relief. I stepped forward, ready to face the beast. 
 "\textcolor{blue}{One last time}, child. Remember when we fought?" \\

{\bf Event:} & $<\!\!|\text{startofcond}|\!\!>$ \textcolor{blue}{two heads} $<\!\!|\text{sepofcond}|\!\!>$ \textcolor{blue}{piercing pain} $<\!\!|\text{sepofcond}|\!\!>$ \textcolor{blue}{near death} $<\!\!|\text{endofcond}|\!\!>$ \\

{\bf Paragraph:} & Then I saw it. The head of the beast was now upon me. It was a creature of such immense power, that it was impossible to tell if it was male or female. It was as if its head was stuck in the air and it had \textcolor{blue}{two heads} to keep from it's eyes. 
 "Who are you?" 
 I turned my head to face the thing that had just taken my blade. The \textcolor{blue}{piercing pain} was just as sharp. 
 "You are a monster. It is \textcolor{blue}{near death}'s door. Be afraid, child. You have fallen for the trap. It will come for you." \\

{\bf Event:} & $<\!\!|\text{startofcond}|\!\!>$ \textcolor{blue}{thousand years ago} $<\!\!|\text{sepofcond}|\!\!>$ \textcolor{blue}{something else} $<\!\!|\text{endofcond}|\!\!>$ \\

{\bf Paragraph:} & It laughed. I could not scream, as I was a \textcolor{blue}{thousand years ago}. I did what I had to do and smiled. It had found a way to stop me. I felt my body tensed. I was ready. 
 "Then who are you?" 
 I did not reply, however. The creature's face had changed and I could feel the tears dripping down its face. 
 "What am I?" 
 "You are the Demon of Excalibur. The one who has given birth to me. I have taken it in order to give you \textcolor{blue}{something else}. I will give you Excalibur. You will not fight me. You are stronger." \\

{\bf Event:} & $<\!\!|\text{startofcond}|\!\!>$ \textcolor{blue}{great power inside} $<\!\!|\text{sepofcond}|\!\!>$ \textcolor{blue}{two heads} $<\!\!|\text{sepofcond}|\!\!>$ \textcolor{blue}{sword back} $<\!\!|\text{endofcond}|\!\!>$ \\

{\bf Paragraph:} & I looked at the \textcolor{blue}{two heads} that were now upon me. The horns on their heads were sharp. Their faces were bright red. I felt a \textcolor{blue}{great power inside} me. I knew what was coming. But I could not help it. I drew my \textcolor{blue}{sword back} and prepared to strike again. 
 "What do you want?" 
 "I want to be a God." 
 "I wish to be a God. So be it." \\

{\bf Event:} & $<\!\!|\text{startofcond}|\!\!>$ \textcolor{blue}{pale hand extended} $<\!\!|\text{sepofcond}|\!\!>$ \textcolor{blue}{demonic face flashed} $<\!\!|\text{sepofcond}|\!\!>$ \textcolor{blue}{looked back} $<\!\!|\text{endofcond}|\!\!>$ \\

{\bf Paragraph:} & I raised my sword and slashed at the horned thing. It was only two of them standing in the room. I raised my sword and grabbed its horns, and in that same instant a \textcolor{blue}{pale hand extended} from its side and grabbed it and sliced it to pieces. I closed my eyes, and \textcolor{blue}{looked back} at the \textcolor{blue}{demonic face flashed}. I closed my eyes and smiled. 
 "Remember when we fought? Remember when we used to live together?" \\

{\bf Event:} & $<\!\!|\text{startofcond}|\!\!>$ \textcolor{blue}{golden sphere shaped like} $<\!\!|\text{sepofcond}|\!\!>$ \textcolor{blue}{old world} $<\!\!|\text{sepofcond}|\!\!>$ \textcolor{blue}{time} $<\!\!|\text{endofcond}|\!\!>$ \\

{\bf Paragraph:} & The demon stared at me for a moment, then turned its head. It had the face of a child, but the horns were a \textcolor{blue}{golden sphere shaped like} a sword, and my blade was a blade that could cut through anything. I looked away for a moment, then let it stare at me for a moment more. 
 "Remember when we used to live together? Remember that \textcolor{blue}{time} we took your father, and then you came to our \textcolor{blue}{old world}
\end{tabular}
\end{tcolorbox}
\caption{Generation examples on test dataset of $\mathtt{WritingPrompts}$ with FIST-$f$ decoding.}
\label{table:example_f_wp}
\end{table*}

\clearpage{} 

\begin{table*}[htbp] \footnotesize{}
\begin{tcolorbox}
\hspace{-7mm}
\begin{tabular}{l p{0.92\linewidth}}

{\bf \textit{Target}} & ---------------------------------------------------------------------------------------------------------------------- \\

{\bf Prompt:} & Watchdogs (Agents of S.H.I.E.L.D.) \\

{\bf Event:} & $<\!\!|\text{startofcond}|\!\!>$ \textcolor{red}{militant terrorist organisation dedicated} $<\!\!|\text{sepofcond}|\!\!>$ \textcolor{red}{improve nitramene weaponry} $<\!\!|\text{sepofcond}|\!\!>$ \textcolor{red}{reluctantly leaves} $<\!\!|\text{endofcond}|\!\!>$ \\

{\bf Paragraph:} & The Watchdogs, a \textcolor{red}{militant terrorist organisation dedicated} to eradicating the Inhumans, destroy an Indiana ATCU facility using nitramene in gel projectiles. Mack, who is in the area visiting his brother Ruben, \textcolor{red}{reluctantly leaves} and joins Daisy and Fitz to investigate, and they inform Coulson of the use of nitramene, leading him to suspect the Watchdogs are being led by former SHIELD agent Felix Blake, who tried to \textcolor{red}{improve nitramene weaponry} in the past. \\

{\bf Event:} & $<\!\!|\text{startofcond}|\!\!>$ \textcolor{red}{recently made redundant} $<\!\!|\text{sepofcond}|\!\!>$ \textcolor{red}{old safe houses} $<\!\!|\text{sepofcond}|\!\!>$ \textcolor{red}{brothers work together} $<\!\!|\text{sepofcond}|\!\!>$ \textcolor{red}{becomes lash permanently} $<\!\!|\text{sepofcond}|\!\!>$ \textcolor{red}{nitramene gel projectile} $<\!\!|\text{endofcond}|\!\!>$ \\

{\bf Paragraph:} & Mack gets into an argument with Ruben, who is unaware that Mack is a SHIELD agent and resents him for barely being present in his life and for helping care for their elderly parents. Ruben was \textcolor{red}{recently made redundant} and is struggling financially, blaming the government for his problems and sympathising with the Watchdogs. Trying to track down Andrew, May enlists Simmons, who suggests that as Andrew is becoming more animalistic, he is running on base instincts and not his usual desires. She also advises against May's intended plan to kill Andrew, believing they can still devise a cure for Terrigenesis before he \textcolor{red}{becomes Lash permanently}. Tracing the Watchdogs' online activity, Daisy and Fitz track down one of their non-radical members and coerce him into revealing the location of the Watchdogs' local headquarters, at a farm. Daisy, Fitz and Mack spy on the farm, but Ruben arrives, intending to apologise to Mack, and blows their cover, forcing the agents to defend him from Watchdogs. In the ensuing fight Fitz is shot with a \textcolor{red}{nitramene gel projectile}, but Daisy stops the implosion by freezing the gel with liquid nitrogen. Following Ruben home, Mack admits he works for SHIELD. They are attacked by five Watchdogs, who saw Daisy's powers in use at the farm but believed Mack to be the Inhuman. The \textcolor{red}{brothers work together} to defend themselves, killing their attackers. Meanwhile, Coulson and Lincoln search one of Blake's \textcolor{red}{old safe houses} and find him there. Blake espouses his hatred of Inhumans and his belief that SHIELD will always use dangerous alien elements to spread terror, and Lincoln tries to kill him with Coulson's permission, only to find 'Blake' is a hologram. \\

{\bf Event:} & $<\!\!|\text{startofcond}|\!\!>$ \textcolor{red}{team later learn} $<\!\!|\text{sepofcond}|\!\!>$ \textcolor{red}{powerful weapons} $<\!\!|\text{endofcond}|\!\!>$ \\

{\bf Paragraph:} & The \textcolor{red}{team later learn} that the Watchdogs were directed to the ATCU facility by Malick, and destroyed the building to cover up the theft of a nuclear weapon, which Blake (a paraplegic after being attacked by Mike Peterson) delivers to Giyera in exchange for more \textcolor{red}{powerful weapons}. \\
\\
{\bf \textit{Generation}} & --------------------------------------------------------------------------------------------------------------------------------- \\

{\bf Prompt:} & Watchdogs (Agents of S.H.I.E.L.D.) \\

{\bf Event:} & $<\!\!|\text{startofcond}|\!\!>$ militant \textcolor{red}{terrorist organisation dedicated} $<\!\!|\text{sepofcond}|\!\!>$ \textcolor{red}{improve nitramene weaponry} $<\!\!|\text{sepofcond}|\!\!>$ \textcolor{red}{reluctantly leaves} $<\!\!|\text{endofcond}|\!\!>$ \\

{\bf Paragraph:} & In Los Angeles, S.Inspector Howerd and dr George Bailey are called to investigate a \textcolor{red}{terrorist organisation dedicated} to gaining control over the world. While trying to \textcolor{red}{improve nitramene weaponry}, the agents are helped by S.Wade who \textcolor{red}{reluctantly leaves} her due to the death of her daughter. While on the rooftop, Bailey gets a call from Amanda who claims the deaths are her fault but S.Wade informs her he has become too attached to her daughter and has decided to destroy the organization. \\

{\bf Event:} & $<\!\!|\text{startofcond}|\!\!>$ \textcolor{red}{recently made redundant} $<\!\!|\text{sepofcond}|\!\!>$ \textcolor{red}{old safe houses} $<\!\!|\text{sepofcond}|\!\!>$ \textcolor{red}{brothers work together} $<\!\!|\text{sepofcond}|\!\!>$ \textcolor{red}{becomes lash permanently} $<\!\!|\text{sepofcond}|\!\!>$ \textcolor{red}{nitramene gel projectile} $<\!\!|\text{endofcond}|\!\!>$ \\

{\bf Paragraph:} & To protect his friend, he is revealed to have been one of the original members of the terrorist organisation, and forced to destroy all of the \textcolor{red}{old safe houses} because his brother was killed during a raid by S.Wade. Bailey and his men battle to bring Amanda to safety. Bailey manages to destroy all the switches in the building, but he is able to gain access to a key which is kept locked. While the others are recovering from their wounds, the agent discovers that the operative is now dead and trapped behind a wall. He \textcolor{red}{becomes lash permanently} and flees in terror. After he manages to break free of her bonds and escape into the building, he destroys all the alarms on the building's computer system. Later, his team of operatives arrive and arrest him. The agents capture him, but not before his \textcolor{red}{brothers work together} to try to stop him. They are able to stop him and he is able to make it out of the building. As the agents prepare to take his place, he escapes into a field where his blood is pouring from his hands. Bailey and his team find him and place him on a grenade in order to destroy his biological weapons. At the same time, the agents stop the production of \textcolor{red}{nitramene gel projectile} from being tested in a \textcolor{red}{recently made redundant}, mutated lab. \\

{\bf Event:} & $<\!\!|\text{startofcond}|\!\!>$ \textcolor{red}{team later learn} $<\!\!|\text{sepofcond}|\!\!>$ powerful \textcolor{red}{weapons} $<\!\!|\text{endofcond}|\!\!>$ \\

{\bf Paragraph:} & As the \textcolor{red}{team later learn} about the \textcolor{red}{weapon}, the agent's body is revealed to be covered in Nitramene gel projectile. Bailey and his team return to the lab to find the agent's mother dead in her bed.
\end{tabular}
\end{tcolorbox}
\caption{Generation examples on test dataset of $\mathtt{WikiPlots}$ with FIST-$a$ decoding.}
\label{table:example_a_wi}
\end{table*}

\clearpage{} 

\begin{table*}[htbp] \footnotesize{}
\begin{tcolorbox}
\hspace{-7mm}
\begin{tabular}{l p{0.92\linewidth}}

{\bf \textit{Target}} & --------------------------------------------------------------------------------------------------------------------------------- \\

{\bf Prompt:} & Watchdogs (Agents of S.H.I.E.L.D.) \\

{\bf Event:} & $<\!\!|\text{startofcond}|\!\!>$ \textcolor{red}{militant terrorist organisation dedicated} $<\!\!|\text{sepofcond}|\!\!>$ \textcolor{red}{improve nitramene weaponry} $<\!\!|\text{sepofcond}|\!\!>$ \textcolor{red}{reluctantly leaves} $<\!\!|\text{endofcond}|\!\!>$ \\

{\bf Paragraph:} & The Watchdogs, a \textcolor{red}{militant terrorist organisation dedicated} to eradicating the Inhumans, destroy an Indiana ATCU facility using nitramene in gel projectiles. Mack, who is in the area visiting his brother Ruben, \textcolor{red}{reluctantly leaves} and joins Daisy and Fitz to investigate, and they inform Coulson of the use of nitramene, leading him to suspect the Watchdogs are being led by former SHIELD agent Felix Blake, who tried to \textcolor{red}{improve nitramene weaponry} in the past. \\

{\bf Event:} & $<\!\!|\text{startofcond}|\!\!>$ \textcolor{red}{recently made redundant} $<\!\!|\text{sepofcond}|\!\!>$ \textcolor{red}{old safe houses} $<\!\!|\text{sepofcond}|\!\!>$ \textcolor{red}{brothers work together} $<\!\!|\text{sepofcond}|\!\!>$ \textcolor{red}{becomes lash permanently} $<\!\!|\text{sepofcond}|\!\!>$ \textcolor{red}{nitramene gel projectile} $<\!\!|\text{endofcond}|\!\!>$ \\

{\bf Paragraph:} & Mack gets into an argument with Ruben, who is unaware that Mack is a SHIELD agent and resents him for barely being present in his life and for helping care for their elderly parents. Ruben was \textcolor{red}{recently made redundant} and is struggling financially, blaming the government for his problems and sympathising with the Watchdogs. Trying to track down Andrew, May enlists Simmons, who suggests that as Andrew is becoming more animalistic, he is running on base instincts and not his usual desires. She also advises against May's intended plan to kill Andrew, believing they can still devise a cure for Terrigenesis before he \textcolor{red}{becomes Lash permanently}. Tracing the Watchdogs' online activity, Daisy and Fitz track down one of their non-radical members and coerce him into revealing the location of the Watchdogs' local headquarters, at a farm. Daisy, Fitz and Mack spy on the farm, but Ruben arrives, intending to apologise to Mack, and blows their cover, forcing the agents to defend him from Watchdogs. In the ensuing fight Fitz is shot with a \textcolor{red}{nitramene gel projectile}, but Daisy stops the implosion by freezing the gel with liquid nitrogen. Following Ruben home, Mack admits he works for SHIELD. They are attacked by five Watchdogs, who saw Daisy's powers in use at the farm but believed Mack to be the Inhuman. The \textcolor{red}{brothers work together} to defend themselves, killing their attackers. Meanwhile, Coulson and Lincoln search one of Blake's \textcolor{red}{old safe houses} and find him there. Blake espouses his hatred of Inhumans and his belief that SHIELD will always use dangerous alien elements to spread terror, and Lincoln tries to kill him with Coulson's permission, only to find 'Blake' is a hologram. \\

{\bf Event:} & $<\!\!|\text{startofcond}|\!\!>$ \textcolor{red}{team later learn} $<\!\!|\text{sepofcond}|\!\!>$ \textcolor{red}{powerful weapons} $<\!\!|\text{endofcond}|\!\!>$ \\

{\bf Paragraph:} & The \textcolor{red}{team later learn} that the Watchdogs were directed to the ATCU facility by Malick, and destroyed the building to cover up the theft of a nuclear weapon, which Blake (a paraplegic after being attacked by Mike Peterson) delivers to Giyera in exchange for more \textcolor{red}{powerful weapons}. \\
\\
{\bf \textit{Generation}} & --------------------------------------------------------------------------------------------------------------------------------- \\

{\bf Prompt:} & Watchdogs (Agents of S.H.I.E.L.D.) \\

{\bf Event:} & $<\!\!|\text{startofcond}|\!\!>$ militant \textcolor{red}{terrorist organisation dedicated} $<\!\!|\text{sepofcond}|\!\!>$ \textcolor{red}{improve nitramene weaponry} $<\!\!|\text{sepofcond}|\!\!>$ \textcolor{red}{reluctantly leaves} $<\!\!|\text{endofcond}|\!\!>$ \\

{\bf Paragraph:} & In Los Angeles, S.Inspector Howerd and dr George Bailey are called to investigate a \textcolor{red}{terrorist organisation dedicated} to gaining control over the world. While trying to \textcolor{red}{improve nitramene weaponry}, the agents are helped by S.Wade who \textcolor{red}{reluctantly leaves} her due to the death of her daughter. While on the rooftop, Bailey gets a call from Amanda who claims the deaths are her fault but S.Wade informs her he has become too attached to her daughter and has decided to destroy the organization. \\

{\bf Event:} & $<\!\!|\text{startofcond}|\!\!>$ \textcolor{red}{team later learn} $<\!\!|\text{sepofcond}|\!\!>$ \textcolor{red}{powerful} weapons $<\!\!|\text{endofcond}|\!\!>$ \\

{\bf Paragraph:} & The \textcolor{red}{team later learn} of Amanda's betrayal, and after an altercation they decide to kill both her and the leader of the group. After discovering the group is still active, they decide to destroy the group while keeping their \textcolor{red}{powers} a secret.
\end{tabular}
\end{tcolorbox}
\caption{Generation examples on test dataset of $\mathtt{WikiPlots}$ with FIST-$r$ decoding.}
\label{table:example_r_wi}
\end{table*}

\clearpage{} 

\begin{table*}[htbp] \footnotesize{}
\begin{tcolorbox}
\hspace{-7mm}
\begin{tabular}{l p{0.92\linewidth}}

{\bf \textit{Target}} & --------------------------------------------------------------------------------------------------------------------------------- \\

{\bf Prompt:} & Watchdogs (Agents of S.H.I.E.L.D.) \\

{\bf Event:} & $<\!\!|\text{startofcond}|\!\!>$ \textcolor{red}{militant terrorist organisation dedicated} $<\!\!|\text{sepofcond}|\!\!>$ \textcolor{red}{improve nitramene weaponry} $<\!\!|\text{sepofcond}|\!\!>$ \textcolor{red}{reluctantly leaves} $<\!\!|\text{endofcond}|\!\!>$ \\

{\bf Paragraph:} & The Watchdogs, a \textcolor{red}{militant terrorist organisation dedicated} to eradicating the Inhumans, destroy an Indiana ATCU facility using nitramene in gel projectiles. Mack, who is in the area visiting his brother Ruben, \textcolor{red}{reluctantly leaves} and joins Daisy and Fitz to investigate, and they inform Coulson of the use of nitramene, leading him to suspect the Watchdogs are being led by former SHIELD agent Felix Blake, who tried to \textcolor{red}{improve nitramene weaponry} in the past. \\

{\bf Event:} & $<\!\!|\text{startofcond}|\!\!>$ \textcolor{red}{recently made redundant} $<\!\!|\text{sepofcond}|\!\!>$ \textcolor{red}{old safe houses} $<\!\!|\text{sepofcond}|\!\!>$ \textcolor{red}{brothers work together} $<\!\!|\text{sepofcond}|\!\!>$ \textcolor{red}{becomes lash permanently} $<\!\!|\text{sepofcond}|\!\!>$ \textcolor{red}{nitramene gel projectile} $<\!\!|\text{endofcond}|\!\!>$ \\

{\bf Paragraph:} & Mack gets into an argument with Ruben, who is unaware that Mack is a SHIELD agent and resents him for barely being present in his life and for helping care for their elderly parents. Ruben was \textcolor{red}{recently made redundant} and is struggling financially, blaming the government for his problems and sympathising with the Watchdogs. Trying to track down Andrew, May enlists Simmons, who suggests that as Andrew is becoming more animalistic, he is running on base instincts and not his usual desires. She also advises against May's intended plan to kill Andrew, believing they can still devise a cure for Terrigenesis before he \textcolor{red}{becomes Lash permanently}. Tracing the Watchdogs' online activity, Daisy and Fitz track down one of their non-radical members and coerce him into revealing the location of the Watchdogs' local headquarters, at a farm. Daisy, Fitz and Mack spy on the farm, but Ruben arrives, intending to apologise to Mack, and blows their cover, forcing the agents to defend him from Watchdogs. In the ensuing fight Fitz is shot with a \textcolor{red}{nitramene gel projectile}, but Daisy stops the implosion by freezing the gel with liquid nitrogen. Following Ruben home, Mack admits he works for SHIELD. They are attacked by five Watchdogs, who saw Daisy's powers in use at the farm but believed Mack to be the Inhuman. The \textcolor{red}{brothers work together} to defend themselves, killing their attackers. Meanwhile, Coulson and Lincoln search one of Blake's \textcolor{red}{old safe houses} and find him there. Blake espouses his hatred of Inhumans and his belief that SHIELD will always use dangerous alien elements to spread terror, and Lincoln tries to kill him with Coulson's permission, only to find 'Blake' is a hologram. \\

{\bf Event:} & $<\!\!|\text{startofcond}|\!\!>$ \textcolor{red}{team later learn} $<\!\!|\text{sepofcond}|\!\!>$ \textcolor{red}{powerful weapons} $<\!\!|\text{endofcond}|\!\!>$ \\

{\bf Paragraph:} & The \textcolor{red}{team later learn} that the Watchdogs were directed to the ATCU facility by Malick, and destroyed the building to cover up the theft of a nuclear weapon, which Blake (a paraplegic after being attacked by Mike Peterson) delivers to Giyera in exchange for more \textcolor{red}{powerful weapons}. \\
\\
{\bf \textit{Generation}} & --------------------------------------------------------------------------------------------------------------------------------- \\

{\bf Prompt:} & Watchdogs (Agents of S.H.I.E.L.D.) \\

{\bf Event:} & $<\!\!|\text{startofcond}|\!\!>$ militant \textcolor{red}{terrorist organisation dedicated} $<\!\!|\text{sepofcond}|\!\!>$ \textcolor{red}{improve nitramene weaponry} $<\!\!|\text{sepofcond}|\!\!>$ \textcolor{red}{reluctantly leaves} $<\!\!|\text{endofcond}|\!\!>$ \\

{\bf Paragraph:} & In Los Angeles, S.Inspector Howerd and dr George Bailey are called to investigate a \textcolor{red}{terrorist organisation dedicated} to gaining control over the world. While trying to \textcolor{red}{improve nitramene weaponry}, the agents are helped by S.Wade who \textcolor{red}{reluctantly leaves} her due to the death of her daughter. While on the rooftop, Bailey gets a call from Amanda who claims the deaths are her fault but S.Wade informs her he has become too attached to her daughter and has decided to destroy the organization. \\

{\bf Event:} & $<\!\!|\text{startofcond}|\!\!>$ \textcolor{blue}{detective dr gilbert lee} $<\!\!|\text{sepofcond}|\!\!>$ \textcolor{blue}{dna report reveals} $<\!\!|\text{endofcond}|\!\!>$ \\

{\bf Paragraph:} & Agent Daryll is dispatched and starts having sessions with \textcolor{blue}{Detective dr Gilbert Lee}. As a result, Lee is able to break into the agency's headquarters and has a breakdown. However, the \textcolor{blue}{DNA report reveals} that Amanda has been murdered and Lee is determined to save Amanda.

\end{tabular}
\end{tcolorbox}
\caption{Generation examples on test dataset of $\mathtt{WikiPlots}$ with FIST-$f$ decoding.}
\label{table:example_f_wi}
\end{table*}

\end{document}